\documentclass{article}

\usepackage[nonatbib, final]{tackling_climate_workshop_style}

\usepackage[utf8]{inputenc} 
\usepackage[T1]{fontenc}    
\usepackage{hyperref}       
\usepackage{url}            
\usepackage{booktabs}       
\usepackage{amsfonts}       
\usepackage{nicefrac}       
\usepackage{microtype}      
\usepackage{graphicx}
\usepackage{amsmath, amssymb}
\usepackage[dvipsnames]{xcolor}
\usepackage{cleveref}
\usepackage[backend=biber, style=nature]{biblatex}
\title{\textit{Sparse Local Implicit Image Function for sub-km Weather Downscaling}}

\author{
  Yago del Valle Inclan Redondo\\
  Recursive Inc.\\
  \texttt{yago.valle@recursiveai.co.jp} \\
  \And
  Enrique Arriaga-Varela \\
  Recursive Inc. \\
  \texttt{enrique.arriaga@recursiveai.co.jp}
  \And
  Dmitry Lyamzin \\
  Recursive Inc. \\
  \texttt{dmitry.lyamzin@recursiveai.co.jp}
  \AND
  Pablo Cervantes \\
  Recursive Inc. \\
  \texttt{pablo.cervantes@recursiveai.co.jp} \\
  \And
  Tiago Ramalho\\
  Recursive Inc. \\
  \texttt{tmramalho@recursiveai.co.jp} \\ 
}

\begin{document}

\maketitle

\begin{abstract}
We introduce SpLIIF to generate implicit neural representations and enable arbitrary downscaling of weather variables. We train a model from sparse weather stations and topography over Japan and evaluate in- and out-of-distribution accuracy predicting temperature and wind, comparing it to both an interpolation baseline and CorrDiff. We find the model to be up to 50\% better than both CorrDiff and the baseline at downscaling temperature, and around 10-20\% better for wind.
\end{abstract}

\section{Introduction}

Accurate, high-resolution weather data is a cornerstone for addressing the multifaceted challenges of global climate change.
\cite{rampal_enhancing_2024, miller_statistical_2025} To meet this need, machine learning provides a powerful alternative to computationally intensive numerical models.\cite{bracco_machine_2025} The field has seen rapid progress, evolving from CNNs,\cite{hohlein_comparative_2020} to GANs,\cite{leinonen_stochastic_2021, oyama_deep_2023} to more recent transformer-based architectures\cite{perez_transformer_2024, sinha_effectiveness_2025, liu_mambads_2024, nguyen_climax_2023} and state-of-the-art diffusion models,\cite{huang_diffda_2024, sundar_taudiff_2025, merizzi_wind_2024, mardani_residual_2025, brenowitz_climate_2025} that excel at generating the realistic, high-frequency spatial details.

Despite these advances, a critical accuracy gap remains, stemming from a reliance on gridded reanalysis datasets. These datasets, while comprehensive, can diverge from the ground-truth weather captured by sparse, irregularly-spaced weather stations.\cite{vanella_comparing_2022} For effective climate adaptation—which demands precise, site-specific forecasting—this discrepancy can be a major hurdle. The key challenge, therefore, is to develop models that can directly leverage the high-fidelity signal of on-the-ground station data to drive more accurate and reliable predictions.

To bridge this gap, we introduce the Sparse Local Implicit Image Function (SpLIIF). While prior work has individually leveraged sparse station data,\cite{liu_kolmogorov_2025, manshausen_generative_2025} topography,\cite{liu_mambads_2024}  or continuous representations,\cite{luo_physics-guided_2025, radhakrishnan_continuous_2024} SpLIIF uniquely combines all three, using an Implicit Neural Representation (INR)\cite{chen_learning_2021} to generate a continuous model of the weather—an approach well-suited to capturing the inherently continuous and non-linear dynamics of atmospheric systems. We compare this approach to a state-of-the-art model (CorrDiff)\cite{mardani_residual_2025} that downscales dense weather maps using a U-Net CNN and generates physically consistent high-resolution results via diffusion. Our approach proves highly accurate and outperforms CorrDiff with a 50\% improvement in temperature prediction and 20\% in wind speed, particularly in complex terrain. Hence, our work contributes to the development of more accurate weather models and therefore more effective and practical tools for global climate change adaptation and resilience.

\section{Methods}

\textbf{Model architecture -- }
SpLIIF can simultaneously process two types of input weather data: dense gridded data $X_d\in \mathbb{R}^{H \times W \times C_{d}}$ (e.g. from numerical weather prediction models), and/or irregularly spaced sparse point data $X_{sp}\in \mathbb{R}^{N \times C_{sp}}$ (e.g. observations from weather stations).

$X_{sp}$ is first transformed into a dense tensor, $X'_{d} \in \mathbb{R}^{H' \times W' \times C_{sp}}$, using an Inverse Distance Weighting (IDW) interpolation scheme with learnable weights. If provided, $X_d$ is interpolated to this intermediate resolution ($H'\times W'$) and concatenated along their channel dimension $\in \mathbb{R}^{H'\times W'\times(C_{sp}+C_{d})}$. This raw feature representation passes through an MLP to project the features into an initial latent space representation $L_0\in \mathbb{R}^{H'\times W'\times C_{L}}$. $L_0$ is then interpolated to the desired resolution and concatenated with the static, high-resolution topography data, $L_1 \in \mathbb{R}^{H'' \times W'' \times (C_{topo}+C_L)}$. This high-resolution latent space is then passed through an Enhanced Deep Super-Resolution (EDSR) network\cite{lim_enhanced_2017} that learns and sharpens spatial features, whilst retaining the size of the feature space.

\textbf{Decoding and loss calculation -- }
From the latent space, output weather variables can be decoded by interpolating the higher-dimensional space at $N$ arbitrary query coordinates before passing through a final MLP that produces the output weather variables $\in \mathbb{R}^{N\times C_{out}}$. When calculating losses, the $N$ locations are positions of weather stations not present in $X_{sp}$ and used to calculate the $L1$ loss (during training) or the RMSE (during evaluation). This decoding step, where the higher-dimensional latent space is interpolated to potentially off-grid station locations is the key aspect that allows SpLIIF to learn a continuous representation of the weather.

\textbf{Datasets -- } \emph{JMA Weather Station Data (sparse input)}\cite{japan_meteorological_agency_amedas_nodate} (training and evaluation), hourly observations of temperature, wind speed and wind direction. Stations are distributed across the entirety of Japan. Median distance between adjacent JMA stations is 0.1 degrees.
\emph{Topography Data} (training and evaluation)\cite{european_space_agency_copernicus_2024}, downsampled to 600m resolution.
\emph{ERA5 (dense input)} (only evaluation)\cite{hersbach_h_era5_2023} just \verb|2m_temperature|, \verb|u10m| and \verb|v10m| variables.

\textbf{Training -- }The model was trained using hourly weather data collected in 2018, from 70\% of the weather stations covering Japan with the exception of Hokkaido. The weather features used were temperature (normalized from [-30, 40]°C to [-1, 1]), wind strength (normalized from [0, 30]m/s to [0, 1]), and wind direction (decomposed along \verb|u| and \verb|v| directions). To manage memory usage and increase batch diversity, this region was partitioned into square patches ($\sim150\times 150$km$\equiv H''=W''=256$). Within each patch, subsets of up to 30 stations are selected, 80\% of which are used as inputs, whilst the rest are used to calculate L1 loss on all variables. Batches comprised 10 patches, exposing the model to a range of topographical features and dynamic weather patterns in each training step, thereby promoting robust generalization. Training took 10h on a single T4 GPU (compared to a few thousand hours on A100s for CorrDiff).

\section{Results}

\subsection{SpLIIF vs Baseline}

\begin{figure}
  \centering
  \includegraphics[width=\textwidth]{"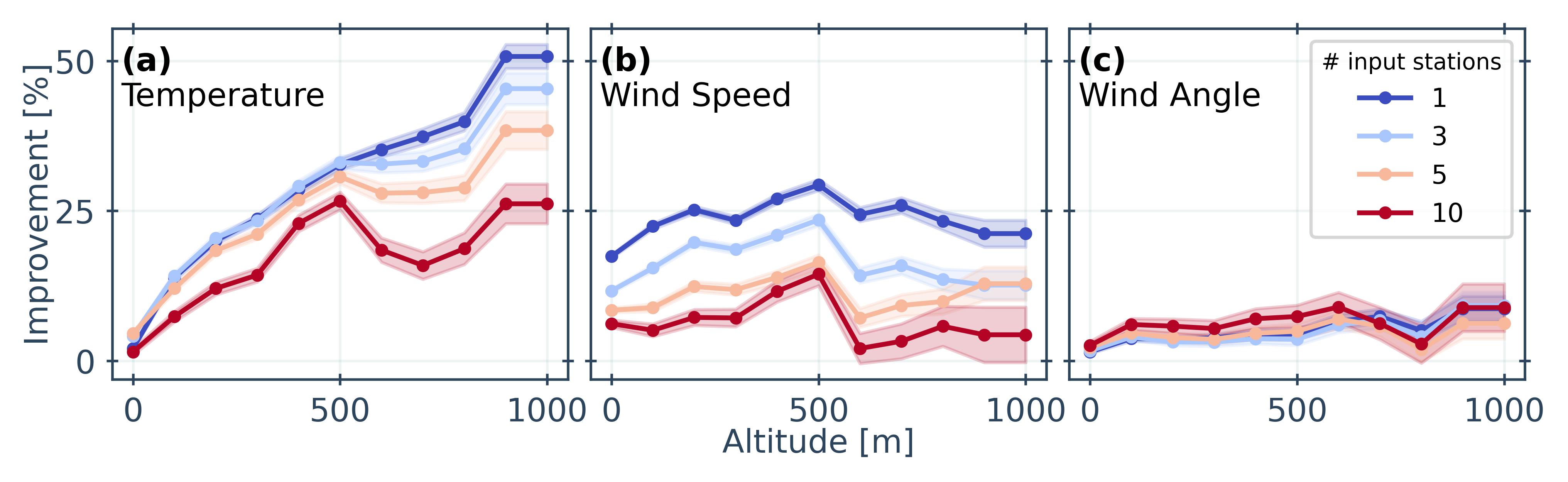"}
  \caption{Percentage improvement in RMSE of SpLIIF over the baseline by altitude and number of input stations for (a) temperature, (b) wind speed, and (c) wind angle. Shaded regions indicate the standard deviation on the mean across the different time slices and spatial patches.}\label{fig_spliif_vs_altitude}
\end{figure}

To evaluate the model's ability to generalize to unseen locations, we measured the Root Mean Square Error (RMSE) between SpLIIF's predictions and ground-truth observations. The evaluation was performed by taking the same input stations as during the training, but comparing the predictions to the 30\% of stations held out from the training set. We used a random 10\% of time slices from the 2018 test year, and performed inference using multiple configurations, varying the number of nearby stations used as input. The results, shown in Figure \ref{fig_spliif_vs_altitude}, are presented as the percentage improvement of SpLIIF's RMSE over an Inverse Distance Weighting (IDW) baseline, and plotted as functions of both the evaluation station's altitude and the number of input stations.

SpLIIF consistently outperformed the baseline across all tested variables, with the magnitude of the improvement depending on the meteorological variable, station altitude, and the number of input stations. For temperature (\Cref{fig_spliif_vs_altitude}a), improvement was most pronounced at higher altitudes. While gains at sea level were modest (1-5\%), they increased significantly with elevation, reaching between 25\% and 50\% at altitudes approaching 1000m.
For wind speed (\Cref{fig_spliif_vs_altitude}b), SpLIIF achieved a 5\% to 30\% improvement over the baseline. Unlike temperature, this improvement did not show a monotonic trend with altitude, but it did increase as the number of input stations decreased. Finally, the improvement for wind angle (\Cref{fig_spliif_vs_altitude}c) was more modest ($\lesssim$10\%), with a slight increase at higher altitudes and no clear dependence on the number of input stations.

These findings can be interpreted by considering the interplay between SpLIIF's two primary data sources: the sparse station observations and the continuous, high-resolution topography map. In data-dense scenarios with many input stations, both model and baseline have sufficient data to construct an accurate downscaled representation. In data-sparse scenarios, however, the model increasingly leverages the topography to inform its predictions, which the baseline has no access to. Hence explaining why the largest performance gain occurs when observational data is limited. Additionally, the more substantial improvements seen for temperature are physically intuitive; temperature has a strong, direct relationship with elevation (lapse rate), which the model can effectively learn from the topography. In contrast, wind patterns are influenced by more complex, local aerodynamic effects that are not as easily captured by elevation alone.

\subsection{SpLIIF vs CorrDiff}

\begin{figure}
  \centering
  \includegraphics[width=\textwidth]{"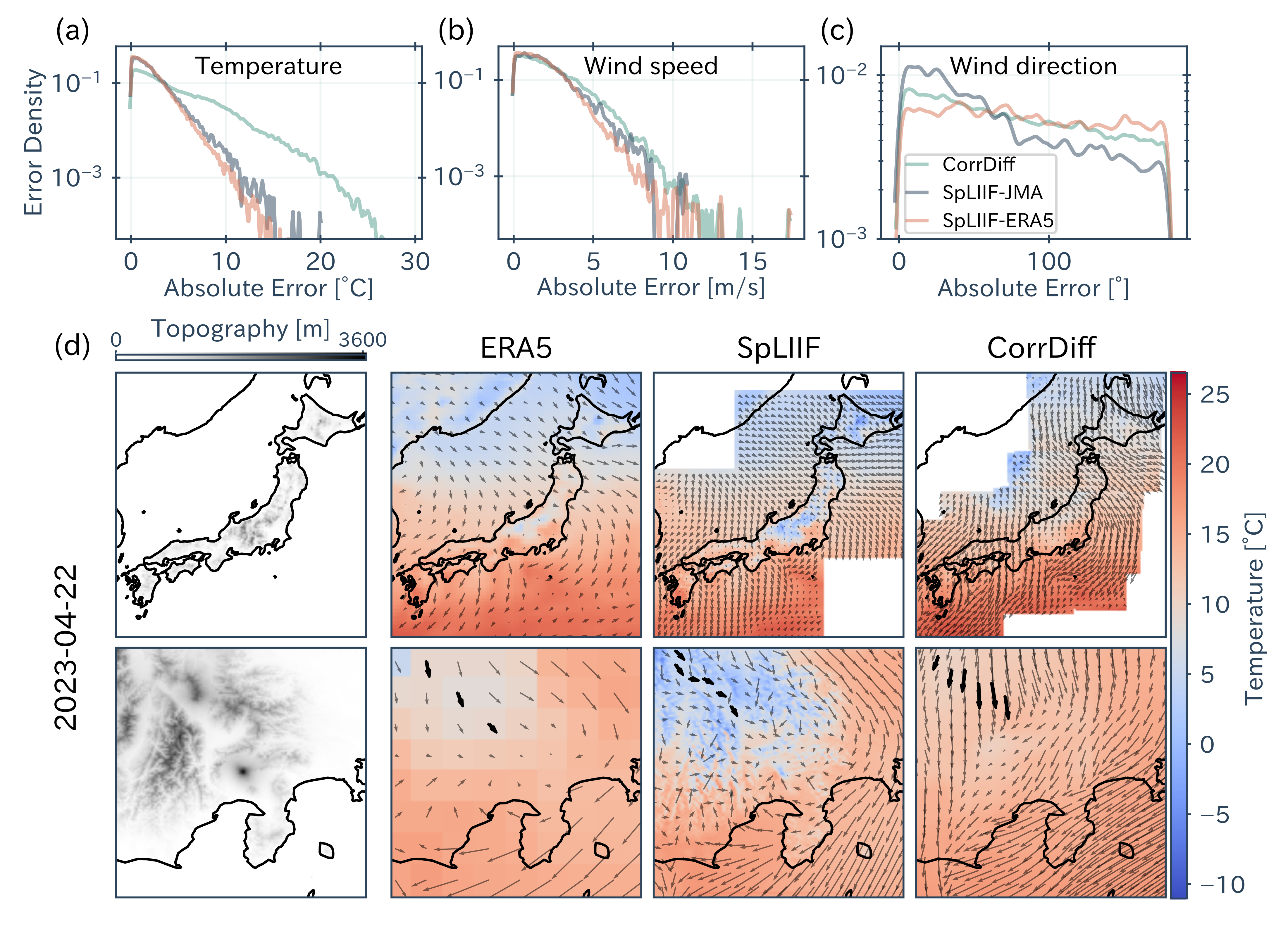"}
  \caption{(a-c) Probability density functions (PDFs) of absolute errors for downscaled predictions CorrDiff, SpLIIF using sparse JMA station data as input (SpLIIF-JMA), and SpLIIF using ERA5 as input (SpLIIF-ERA5). Error distributions are shown for (a) temperature (°C), (b) wind speed (m/s), and (c) wind direction (°).
    (d) Example inference during a cold front. First column shows topography, and the next three columns respectively show the temperature and wind from ERA5, SpLIIF-ERA5 and CorrDiff. Bolded arrows are a guide-to-the-eye for wind along a valley.
  }\label{fig_spliif_vs_corrdiff}
\end{figure}

Next, we evaluated SpLIIF's out-of-distribution generalization against a state-of-the-art baseline, CorrDiff. We compared the absolute error of both models on data from Hokkaido, a region unseen during training for either model, using ground-truth data from JMA stations for 10\% of hours in 2023. For SpLIIF, we tested inference using both JMA station data and ERA5 reanalysis as inputs. \Cref{fig_spliif_vs_corrdiff} plots the error distributions, where superior performance is indicated by a higher density of small errors and a lower density of large errors.

For temperature (\Cref{fig_spliif_vs_corrdiff}a), SpLIIF demonstrates a clear advantage, producing approximately twice as many predictions with an absolute error below 1°C and a tenfold reduction in errors exceeding 10°C, leading to a 50\% reduction in the mean error. Notably, SpLIIF's performance was even stronger when downscaling from ERA5 inputs, despite being trained exclusively on JMA station data. This result highlights the INR architecture's ability to generalize to different input data sources without re-training. The performance gains for wind speed were more moderate, with SpLIIF showing 15\% more low-error predictions and a threefold reduction in high-error predictions (\Cref{fig_spliif_vs_corrdiff}b), leading to a 20\% reduction in the mean error. For wind direction (\Cref{fig_spliif_vs_corrdiff}c), SpLIIF using JMA inputs yielded the best results, with a 10\% reduction in the mean error. We hypothesize that direct ground-truth observations provide a crucial signal for local wind dynamics—influenced by microscale topography—that is absent in coarser reanalysis data.

A qualitative comparison further highlights these differences. \Cref{fig_spliif_vs_corrdiff}d shows a downscaled cold front from April 2023. SpLIIF's temperature map contains more high-resolution features that directly correspond to the underlying topography, details that are absent in the CorrDiff output. Similarly, SpLIIF's wind fields show more complex patterns in mountainous regions, with wind directions aligning more closely with valley structures and contouring mountain ranges, demonstrating a superior ability to capture topographically-induced weather effects.

\section{Conclusion and future work}

In this work, we introduced SpLIIF, a downscaling model that combines a continuous implicit neural representation with sparse station data and high-resolution topography. Our model significantly outperforms traditional baselines, reducing temperature RMSE by up to 50\% in complex terrain, and surpasses a state-of-the-art diffusion model in out-of-distribution tests. These results highlight the value of integrating real-world, sparse data and demonstrate that INR-based architectures offer a flexible and powerful alternative to the grid-dependent backbones common in generative models. By generating more physically plausible weather fields from sparse inputs, SpLIIF contributes to the development of more effective tools for climate change adaptation.

Future work will enhance SpLIIF's applicability to climate action by extending it to predict variables critical for extreme events, like precipitation, and by exploring hybrid architectures that integrate our data-driven approach with physically-constrained models.

\printbibliography

\end{document}